\documentclass[10pt,twocolumn,letterpaper]{article}

\usepackage[T1]{fontenc}
\usepackage{inconsolata}
\usepackage{cvpr}
\usepackage{times}
\usepackage{epsfig}
\usepackage{graphicx}
\usepackage{amsmath}
\usepackage{amssymb}
\usepackage{multirow}
\usepackage{booktabs}
\usepackage{cellspace}
\usepackage{subcaption}
\usepackage{array}
\usepackage[hang,flushmargin]{footmisc}
\usepackage[table,usenames,dvipsnames]{xcolor}
\usepackage{collcell}
\usepackage{pgfplots}
\usepackage{cite}
\pgfplotsset{compat=newest}
\usepgfplotslibrary{colormaps}

\usepackage[breaklinks=true,bookmarks=false]{hyperref}
\usepackage[capitalise]{cleveref}
\cvprfinalcopy

\def\cvprPaperID{****} 

\makeatletter
\renewcommand{\paragraph}{%
  \@startsection{paragraph}{4}%
  {\z@}{0.3ex \@plus 1ex \@minus .2ex}{-1em}%
  {\normalfont\normalsize\bfseries}%
}
\makeatother

\newcommand{\midsepremove}{\aboverulesep=0mm \belowrulesep=0mm}%
\newcommand{\midsepdefault}{\aboverulesep=0.605mm \belowrulesep=0.984mm}%

\midsepremove%
\definecolor{ao (english)}{rgb}{0.0, 0.5, 0.0}
\definecolor{amaranth}{rgb}{0.9, 0.17, 0.31}
\definecolor{green}{rgb}{0.34425221068819684, 0.608073817762399, 0.15647827758554403}
\definecolor{pink}{rgb}{0.7975394079200308, 0.19784698193002692, 0.5391772395232602}

\newcommand{\ci}[1]{\textcolor{green}{#1}}
\newcommand{\cd}[1]{\textcolor{pink}{#1}}

\begin{document}

\title{An Evaluation of Action Recognition Models on EPIC-Kitchens}

\author{Will Price\\
University of Bristol, UK\\
{\tt\small will.price@bristol.ac.uk}
\and
Dima Damen\\
University of Bristol, UK\\
{\tt\small dima.damen@bristol.ac.uk}
}

\maketitle

\section{Introduction}
\noindent We benchmark contemporary action recognition models (TSN~\cite{wang2018_TemporalSegmentNetworks}, TRN~\cite{zhou2017_TemporalRelationalReasoning}, and TSM~\cite{lin2018_TSMTemporalShift}) on the recently introduced EPIC-Kitchens dataset~\cite{damen2018_ScaleingEgocentricVision} and release pretrained models on GitHub\footnote{\href{https://github.com/epic-kitchens/action-models}{\texttt{github.com/epic-kitchens/action-models}}} for others to build upon.
In contrast to popular action recognition datasets like Kinetics~\cite{kay2017_KineticsHumanAction}, Something-Something~\cite{goyal2017_SomethingSomethingVideo}, UCF101~\cite{soomro2012_UCF101DatasetHuman}, and HMDB51~\cite{kuehne2011_HMDBLargeVideo}, EPIC-Kitchens is shot from an egocentric perspective and captures daily actions in-situ.
In this report, we aim to understand how well these models can tackle the challenges present in this dataset, such as its long tail class distribution, unseen environment test set, and multiple tasks (verb, noun and, action classification).
We discuss the models' shortcomings and avenues for future research.

\section{Models}

\noindent We benchmark 3 models: Temporal Segment Networks~(TSN)~\cite{wang2018_TemporalSegmentNetworks}, Temporal Relational Networks (TRN)~\cite{zhou2017_TemporalRelationalReasoning}, and Temporal Shift Module (TSM) based networks~\cite{lin2018_TSMTemporalShift}, including a variety of their variants.
These models are evaluated under a uniform training and testing regime to ensure the results are directly comparable.
TSN is the earliest model of the three and both TRN and TSM can be viewed as evolutionary descendants of TSN, integrating temporal modelling.
In the following paragraphs, we provide an explanation of how network inputs are sampled and a brief summary of the design of each network.

\paragraph{Sampling} Inputs to the models, \textit{snippets}, are sampled according to the TSN sampling strategy. An action clip is split into $n$ equally sized \textit{segments} and a snippet is sampled at a random position within each of these.
For an RGB network, the input is a single frame and for a flow network it is a stack of 5 $(u,v)$ optical flow pairs (proposed in the two-stream CNN~\cite{simonyan2014_TwoStreamConvolutional}).

\paragraph{TSN~\cite{wang2018_TemporalSegmentNetworks}} Temporal Segment Networks propagate each snippet through a 2D CNN backbone and aggregate the class scores across segments through average or max pooling.
As a consequence, TSN is unable to learn temporal correlations across segments.
TSN is typically trained on RGB and optical flow modalities and combined by late-fusion.

\paragraph{TRN)~\cite{zhou2017_TemporalRelationalReasoning}} Temporal Relation Networks propagate snippets through a 2D CNN, like in TSN, up to the pre-classification layer. These produce features rather than class confidence scores.
In order to support inter-segment temporal modelling, these segment-level features are then processed by a modified relational module~\cite{santoro2017_SimpleNeuralNetwork} sensitive to item ordering.
Two variants of the TRN module exists: a single scale version which computes a single $n$-segment relation, and a multi-scale (M-TRN) variant which computes relations over ordered sets of segment features of size 2 to $n$.
Once the relational features have been computed, they are summed and fed to a classification layer.

\paragraph{TSM~\cite{lin2018_TSMTemporalShift}} These networks functionally operate just like TSN, snippets are sampled per segment, propagated through the backbone, and then averaged.
However, unlike TSN, the backbone is modified to support reasoning across segments by shifting a proportion of the filter responses across the temporal dimension.
This opens the possibility for subsequent convolutional layers to learn temporal correlations.

\section{Experiments}

\noindent In this section, we examine how a variety of factors impact model performance such as backbone choice, input modality, and temporal support.
We analyse model performance across tasks from the perspective of the more defining characteristics of the dataset: the long-tail class distribution, and the domain gap between the seen and unseen kitchen test sets.

\subsection{Experimental details}
\paragraph{Tasks} EPIC-Kitchens has three tasks within the action recognition challenge: classifying the verb, noun, and action (the verb-noun pair) of a given trimmed video.
We follow the approach in~\cite{damen2018_ScaleingEgocentricVision}, and  replace the classification layer of each model with two output FC layers, one for verbs $v$ and one for nouns $n$.
The models are trained with an averaged softmax cross-entropy loss over each classification layer: $\mathcal{L} = 0.5 (\mathcal{L}_n + \mathcal{L}_v)$.
We obtain action predictions from verb and noun predictions assuming the tasks are independent.
Later, we examine the impact of integrating action priors computed from the training set for action classification.
Performance on these tasks are evaluated on two test sets: seen kitchens (S1) and unseen kitchens (S2).
The unseen kitchens test set contains videos from novel environments, whereas the seen kitchens split contains videos from the same environments used in training.




\definecolor{lowcolor}{rgb}{0.5543118699137254, 0.6900970112156862, 0.9955155482352941}
\definecolor{centralcolor}{rgb}{0.8674276350862745, 0.864376599772549, 0.8626024620196079}
\definecolor{highcolor}{rgb}{0.9566532109764706, 0.598033822717647, 0.4773022923529412}

\definecolor{lowcolor}{rgb}{0.9085736255286428, 0.5926182237600922, 0.770319108035371}
\definecolor{centralcolor}{rgb}{0.9673202614379085, 0.968473663975394, 0.9656286043829296}
\definecolor{highcolor}{rgb}{0.6032295271049597, 0.8055363321799309, 0.3822376009227222}
\pgfplotsset{%
  colormap={PiYG}{%
    rgb=(0.9085736255286428, 0.5926182237600922, 0.770319108035371)%
    rgb=(0.9673202614379085, 0.968473663975394, 0.9656286043829296)%
    rgb=(0.6032295271049597, 0.8055363321799309, 0.3822376009227222)%
  }%
}

\def\pgfplotsshowcolormap#1{%
    \pgfplotscolormapifdefined{#1}{\relax}{%
        \pgfplotsset{colormap/#1}%
    }%
    \pgfplotscolormaptoshadingspec{#1}{1cm}\result
    \def\tempb{\pgfdeclarehorizontalshading{tempshading}{0.2cm}}%
    \expandafter\tempb\expandafter{\result}%
    \pgfuseshading{tempshading}%
}

\newlength{\savedtabcolsep}
\setlength{\savedtabcolsep}{\tabcolsep}
\setlength{\tabcolsep}{1ex}

\newcommand{\ApplyGradient}[1]{%
  \pgfmathsetmacro{\Val}{#1}%
  \pgfmathsetmacro{\MiddleVal}{\MinVal + (\MaxVal - \MinVal) / 2}%
  \ifdim #1 pt > \MiddleVal pt%
      \pgfmathsetmacro{\PercentColor}{max(min(100.0*(#1 - \MiddleVal)/(\MaxVal-\MiddleVal),100.0),0.00)}%
      \edef\HeatCell{\noexpand\cellcolor{highcolor!\PercentColor!centralcolor}}%
      \HeatCell$#1$%
  \else
      \pgfmathsetmacro{\PercentColor}{max(min(100.0*(\MiddleVal - \Val)/(\MiddleVal-\MinVal),100.0),0.00)} %
      \edef\HeatCell{\noexpand\cellcolor{lowcolor!\PercentColor!centralcolor}}%
      \HeatCell$#1$%
  \fi%
}

\newcolumntype{\C}[2]{>{\def\MinVal{#1}\def\MaxVal{#2}\collectcell\ApplyGradient}c<{\endcollectcell}}

\begin{table*}[t]
  \small
  \centering
  \midsepremove
  \begin{tabular}{lll %
    \C{47.97}{62.68}\C{36.46}{52.27}\C{83.87}{88.85}\C{73.64}{79.55}
    \C{23.19}{41.88}\C{18.50}{26.02}\C{47.02}{66.43}\C{40.25}{49.47}
    \C{16.03}{29.90}\C{11.30}{17.79}\C{32.92}{49.81}\C{26.08}{32.67}
    }
    \toprule
                                                          &                        &          & \multicolumn{4}{Sc}{Verb}  &  \multicolumn{4}{Sc}{Noun} & \multicolumn{4}{Sc}{Action} \\
                                                                                              \cmidrule(lr){4-7}          \cmidrule(lr){8-11}           \cmidrule(l){12-15}
                                                          &                        &          & \multicolumn{2}{c}{Top-1} & \multicolumn{2}{c}{Top-5} & \multicolumn{2}{c}{Top-1} & \multicolumn{2}{c}{Top-5} & \multicolumn{2}{c}{Top-1} & \multicolumn{2}{c}{Top-5} \\
                                                                                              \cmidrule(lr){4-5}          \cmidrule(lr){6-7}          \cmidrule(lr){8-9}          \cmidrule(lr){10-11}         \cmidrule(lr){12-13}        \cmidrule(lr){14-15}
    BB                                                    & Model                  & Modality & \multicolumn{1}{c}{S1}     & \multicolumn{1}{c}{S2}    & \multicolumn{1}{c}{S1}     & \multicolumn{1}{c}{S2}      & \multicolumn{1}{c}{S1}     & \multicolumn{1}{c}{S2}    & \multicolumn{1}{c}{S1}     & \multicolumn{1}{c}{S2}       & \multicolumn{1}{c}{S1}     & \multicolumn{1}{c}{S2}    & \multicolumn{1}{c}{S1}     & \multicolumn{1}{c}{S2} \\
    \midrule
  \multirow{10}{*}{\rotatebox[origin=c]{90}{BN-Inception}} & \multirow{3}{*}{TSN}  & RGB      & 47.97 & 36.46 & 87.03 & 74.36   & 38.85 & 22.64 & 65.54 & 46.94    & 22.39 & 11.30 & 44.75 & 26.32 \\ 
                                                          &                        & Flow     & 51.68 & 47.35 & 84.63 & 76.95   & 26.82 & 21.20 & 50.64 & 42.47    & 16.76 & 13.49 & 33.75 & 27.52 \\ 
                                                          &                        & Fusion   & 54.70 & 46.06 & 87.24 & 76.65   & 40.11 & 24.27 & 65.81 & 49.27    & 25.43 & 14.78 & 45.69 & 29.81 \\ 
                                                          \cmidrule{2-15}
                                                          & \multirow{3}{*}{TRN}   & RGB      & 58.26 & 47.29 & 87.14 & 76.54   & 36.32 & 22.91 & 63.30 & 44.73    & 25.46 & 15.06 & 45.66 & 28.99 \\ 
                                                          &                        & Flow     & 55.20 & 50.32 & 84.04 & 77.67   & 23.95 & 19.02 & 47.02 & 40.25    & 16.03 & 12.77 & 32.92 & 27.62 \\ 
                                                          &                        & Fusion   & 61.04 & 51.83 & 87.46 & 79.11   & 37.90 & 24.75 & 63.69 & 47.35    & 26.54 & 16.59 & 46.37 & 31.14 \\ 
                                                          \cmidrule{2-15}
                                                          & \multirow{3}{*}{M-TRN} & RGB      & 57.66 & 45.41 & 86.91 & 76.34   & 37.94 & 23.90 & 63.78 & 46.33    & 26.62 & 15.57 & 46.39 & 29.57 \\ 
                                                          &                        & Flow     & 55.92 & 51.38 & 84.44 & 77.74   & 24.88 & 20.69 & 48.37 & 40.83    & 16.78 & 14.00 & 34.09 & 28.75 \\ 
                                                          &                        & Fusion   & 61.12 & 51.62 & 87.71 & 78.42   & 39.28 & 26.02 & 64.36 & 48.99    & 27.86 & 17.34 & 47.56 & 32.57 \\ 
    \midrule
    \multirow{13}{*}{\rotatebox[origin=c]{90}{ResNet-50}} & \multirow{3}{*}{TSN}   & RGB      & 49.71 & 36.70 & 87.19 & 73.64   & 39.85 & 23.11 & 65.93 & 44.73    & 23.97 & 12.77 & 46.14 & 26.08 \\ 
                                                          &                        & Flow     & 53.14 & 47.56 & 84.88 & 76.89   & 27.76 & 20.28 & 51.29 & 42.23    & 18.03 & 13.11 & 35.18 & 27.83 \\ 
                                                          &                        & Fusion   & 55.50 & 45.75 & 87.85 & 77.40   & 41.28 & 25.13 & 66.53 & 48.11    & 26.89 & 15.40 & 47.35 & 30.01 \\ 
                                                          \cmidrule{2-15}
                                                          & \multirow{3}{*}{TRN}   & RGB      & 58.82 & 47.32 & 86.60 & 76.92   & 37.27 & 23.69 & 62.96 & 46.02    & 26.62 & 15.71 & 46.09 & 30.01 \\ 
                                                          &                        & Flow     & 55.16 & 50.39 & 83.87 & 77.71   & 23.19 & 18.50 & 47.33 & 40.70    & 15.77 & 12.02 & 33.08 & 27.42 \\ 
                                                          &                        & Fusion   & 61.60 & 52.27 & 87.20 & 79.55   & 38.41 & 25.74 & 63.37 & 47.87    & 27.58 & 17.79 & 46.44 & 32.20 \\ 
                                                          \cmidrule{2-15}
                                                          & \multirow{3}{*}{M-TRN} & RGB      & 60.16 & 46.94 & 87.18 & 75.21   & 38.36 & 24.41 & 64.67 & 46.71    & 28.23 & 16.32 & 47.89 & 29.74 \\ 
                                                          &                        & Flow     & 56.79 & 50.36 & 84.91 & 77.67   & 25.00 & 20.28 & 48.70 & 41.45    & 17.24 & 13.42 & 34.80 & 29.02 \\ 
                                                          &                        & Fusion   & 62.68 & 52.03 & 87.96 & 78.90   & 39.82 & 25.88 & 64.94 & 49.03    & 29.41 & 17.86 & 48.91 & 32.54 \\ 
                                                          \cmidrule{2-15}
                                                          & \multirow{3}{*}{TSM}   & RGB      & 57.88 & 43.50 & 87.14 & 73.85   & 40.84 & 23.32 & 66.10 & 46.02    & 28.22 & 14.99 & 49.12 & 28.06 \\ 
                                                          &                        & Flow     & 58.08 & 52.68 & 85.88 & 79.11   & 27.49 & 20.83 & 50.27 & 43.70    & 19.14 & 14.27 & 36.90 & 29.60 \\ 
                                                          &                        & Fusion   & 62.37 & 51.96 & 88.55 & 79.21   & 41.88 & 25.61 & 66.43 & 49.47    & 29.90 & 17.38 & 49.81 & 32.67 \\ 
    \bottomrule
  \end{tabular}
  \caption[]{Backbone (BB) comparison using 8 segments in both training and
    testing evaluating top-1/5 accuracy across tasks.
    S1 denotes the seen test set, and S2 the unseen test set.
    Cells are coloured on a per column basis: \textcolor{pink}{low}
    \begin{tikzpicture}%
      \pgfplotscolorbardrawstandalone[%
        colormap name=PiYG,%
        colorbar horizontal,%
        colorbar style={%
          height=0.18cm,%
          width=2cm,%
          hide axis,%
        }%
      ]%
    \end{tikzpicture} \textcolor{green}{high}. }
  \label{tab:backbone-comparison}
  \midsepdefault
\end{table*}
\setlength{\tabcolsep}{\savedtabcolsep}

\paragraph{Training}
We train all models with a batch size of 64 for 80 epochs using an ImageNet pretrained model for initialisation.
SGD is used for optimisation with momentum of 0.9.
A weight decay of $5\times 10^{-4}$ is applied and gradients are clipped at 20.
We replace the backbone's classification layer with a dropout layer, setting $p = 0.7$.
We train RGB models with an initial learning rate (LR) of 0.01 for ResNet-50 based models and 0.001 for BN-Inception models.
flow models are trained with an LR of 0.001.
These LRs were the maximum we could achieve whilst maintaining convergence.
The LR is decayed by a factor of 10 at epochs 20 and 40.
\paragraph{Testing}
Models are evaluated using 10 crops (center and corner crops as well as their horizontal flips) for each clip.
The scores from these are averaged pre-softmax to produce a single clip-level score.
Fusion results are obtained by averaging the softmaxed scores obtained for each modality.

\subsection{Results}

\paragraph{Backbone choice}
To choose a high performing backbone, we compare BN-Inception~\cite{ioffe2015_BatchNormalizationAccelerating,szegedy2015_GoingDeeperConvolutions} to ResNet-50~\cite{he2016_DeepResidualLearning} across the 3 models, training and testing with 8 segments.
We did not test TSM with BN-Inception as the authors state that the shift module is harmful unless placed in a residual branch~\cite{lin2018_TSMTemporalShift}.
The top-1/5 accuracy across tasks is reported in \cref{tab:backbone-comparison} where the results show ResNet-50 to be superior to BN-Inception in 14/18 cases when examining top-1 action accuracy across both test sets.

\paragraph{Aggregate performance}
We now compare models with ResNet-50 backbones across tasks in \cref{tab:backbone-comparison} using top-1/5 accuracy.
On the verb task, an intrinsically more temporal problem than classifying nouns, both M-TRN and TSM out perform TSN, especially when operating on RGB frames instead of flow.
This can be explained by TSN's inability to learn inter-segment correlations as only average or max pooling is used in aggregating class scores across segments.
TSN flow models outperform their RGB counterparts; this can be attributed to the network being passed temporal information in the form of stacked optical flow frames.
The 2D convolutions inside the network can learn temporal relations within the stack. 
Both \mbox{(M-)}TRN and TSM flow models outperform TSN flow showing that inter-segment reasoning is complimentary to intra-segment reasoning.

Unlike verb classification, noun classification does not rely on temporal modelling \textit{as much} since objects can be recognised from a single frame.
TSM and TSN perform best on this task, with TRN models lagging 2--3\% points behind.
A possible explanation for the observed drop is that the relation module within TRN places heavy emphasis on extracting temporal relational information, which is of little relevance in recognising objects.
Noun performance drops considerably across models when switching from RGB to flow as the former is a much better modality for recognising objects.
Unexpectedly, TSM improves top-1 noun accuracy by 1\% point over TSN.
Additionally, we find all fusion models improve over the RGB models alone.
We hypothesise that the temporal information here is helping disambiguate the action relevant object from those that are simply present in the environment.

Classifying actions, the joint task of classifying both verb and noun, is clearly very challenging, with the best top-1 accuracy on actions being 29.9\% and 17.9\% for the seen and unseen test set respectively.
Even at top-5, the best results are 49.8\% and 32.8\%.
Despite flow's superior results on verb classification on the unseen test set, the inferior noun performance drags flow models below RGB models on both test sets. 

An enduring approach, pioneered by the 2SCNN~\cite{simonyan2014_TwoStreamConvolutional}, has been to ensemble networks trained on different modalities through late fusion at test-time.
Averaged across all model variants, fusing both modalities results in a 2.9\%, 5.8\%, and 9.7\% relative improvement over the best performing single modality model for verb, noun, and action classification respectively.
The best model on the seen test set is TSM fusion, followed by M-TRN fusion. On the unseen test set, the trend is reversed with M-TRN out-performing TSM.

\paragraph{Novel environment robustness}
It is interesting to examine the relative drop in model performance from the seen to unseen test set to determine the models' ability to generalise to new environments.
\cref{tab:backbone-comparison} shows that flow models are more robust to the domain gap between the seen kitchens and unseen kitchens test sets only suffering an average 22\% relative drop in top-1 action accuracy compared to a 44\% drop for RGB models, and 39\% for fused models.
The domain gap on fused models suggests that the RGB model's predictions dominates those of the flow model.
We find that flow models consistently outperform RGB models for verb classification on the unseen test set.
We hypothesis this is due to the absence of appearance information in optical flow, forcing flow models to focus on motion.
Motion is more environment-invariant and salient to the classification of verbs than the visual cues the RGB models will use.

\begin{figure*}[p]






  \includegraphics[height=0.93\textheight]{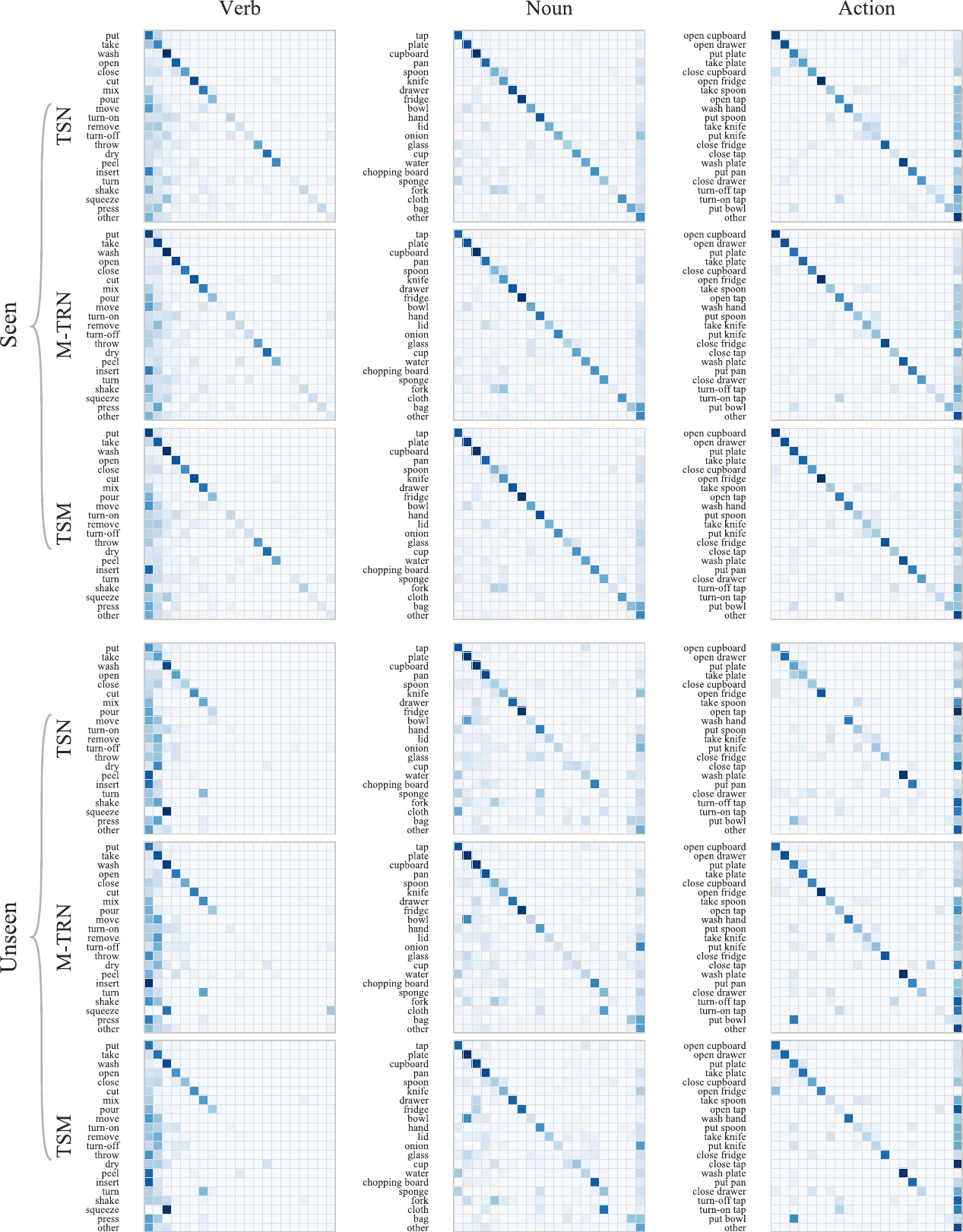}
  \caption{Fusion models' performance on top-20 most frequent classes in
    training. Classes are ordered from top to bottom in descending order of
    frequency and any classes outside the top-20 are grouped into a super-class
    labelled `other'. [Best viewed on screen]}
  \label{fig:long-tail-confusion-matrices}
\end{figure*}

\paragraph{Class performance analysis}
To further understand the differences between models, we look at confusion amongst the top-20 most frequent classes in training in \cref{fig:long-tail-confusion-matrices}.

The verb classification results show the top-3 verbs (accounting for 53\% of the actions in training) dominate predictions due to the dataset imbalance, with this effect being especially pronounced in the unseen test set.
Classes outside the top-20 are rarely correctly classified and instead are classified into one of the majority classes.
The fine-grained nature of the verbs seems to pose challenges, particularly in the unseen test set, with similar classes being confused, such as `move' with `put'/`take', `turn' with `mix', and `insert' with `put'.
TSN shows increased confusion between classes that differ primarily in their temporal aspects (e.g. `put' vs `take'), compared to TSM and TRN.
The generic class `move' is hardest to classify, for all models.

For noun classification, the confusion matrices show the models don't struggle as much to classify less frequent classes compared to verb classification. 
This is likely as a result of the models benefiting from pretraining on the large-scale ImageNet dataset.
However, when fine-tuned, some overfitting to seen environments is observed, as the unseen test set matrices demonstrate that the models generalise less well to new objects.
Like the verb results, the fine-grained classes pose a challenge with confusion between similar objects like `fork' with `spoon', and `bowl' with `plate' occurring.
Another interesting contrast between the verb and noun tasks is that the top-20 verbs almost never get misclassified into any of the classes outside the top-20, whereas for nouns, there are more misclassifications of top-20 nouns into the long-tail.

\newlength{\cnfmatrixsep}

\begin{figure*}[ht]
  \centering
  \includegraphics[width=0.8\textwidth]{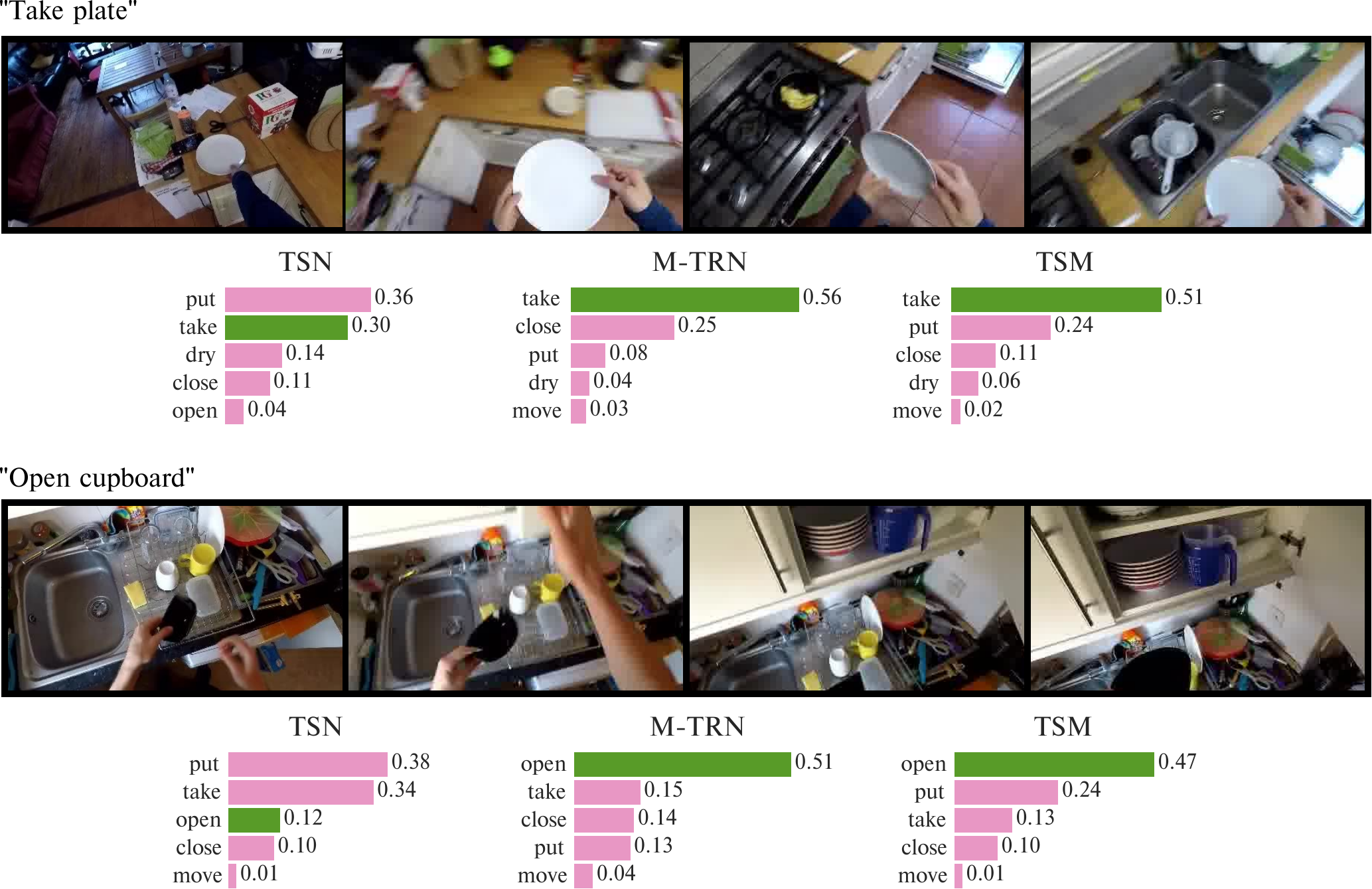}
  \caption{Two examples demonstrating where models capable of temporal reasoning, TRN and TSM,
    improve over TSN. The bar charts show the model's scores on the above
    example with the correct class' score shown in \textcolor{green}{green}.}
  \label{fig:qualitative-results}
\end{figure*}

For action classification, the models perform well on frequent actions, but suffer more misclassifications into the long tail than nouns (as evidenced by the confusion into `other' classes).
Confusion within the top-20 actions highlight an issue not visible from the verb and noun matrices: semantically identical classes like `turn-on tap' are confused with `open tap'.
Whilst these are different classes in the dataset, they refer to the same action.
This highlights an issue with the open vocabulary annotation process employed by the dataset: annotators may use different phrases for describing the same action.

We provide qualitative examples in \cref{fig:qualitative-results} where TSM and M-TRN correctly classify the actions, but TSN fails.
In the top example TSN confuses `put' and `take' as a result of averaging the scores across segments, and thus discarding temporal ordering.
M-TRN and TSM show a much larger disparity between the scores of these classes indicating they have better learnt the difference.
In the bottom example, TSN again struggles to correctly classify the action.
The temporal bounds are quite wide and capture frames just after someone has picked up a bowl, they then open the cupboard and are about to place the bowl.
M-TRN and TSM, through their ability to draw correlations across segments, are able to disambiguate the correct class from the action which came before and comes after.

\begin{figure*}[ht]
  \centering
  \includegraphics{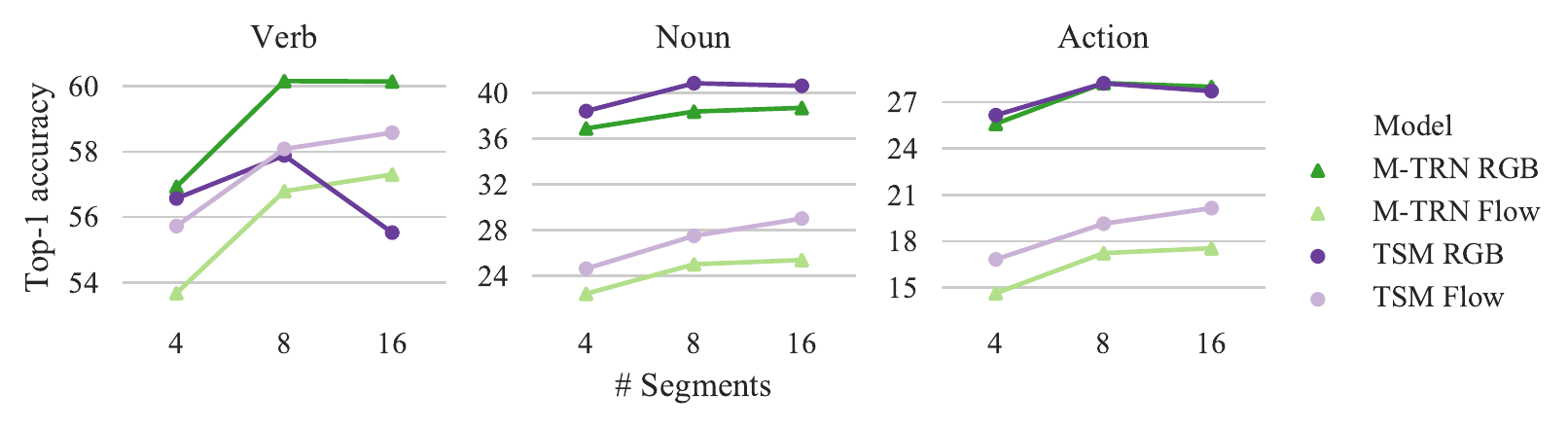}%
  \vspace{-6pt}
  \caption{Top-1 accuracy on the seen test set when varying number of segments (during both training/testing) for M-TRN and TSM.}
  \label{fig:temporal-support-performance}
\end{figure*}

\paragraph{Temporal support}
How many frames/optical flow snippets does the network need to see before performance saturates?
We examine the answer to this question by training models with different numbers of segments, presenting results in \cref{fig:temporal-support-performance}.
Overall, flow models benefit more from increasing temporal support, showing monotonically increasing performance, unlike RGB models whose performance saturates at 8 frames, even dropping for the action task when using 16 frames.
Curiously, the RGB TSM model is severely harmed by using 16 segments instead of 8, unlike its flow counterpart whose performance improves moving from 8 to 16 segments.
This is in contrast to the authors results on Kinetics and Something-something which show an improvement in using 16 frames over 8.
This drop was consistently observed across varying LRs suggesting this is not due to a suboptimal learning rate.

\paragraph{Action priors}
In the previous sections, action predictions have been computed assuming independence between verbs and nouns
\begin{equation}
  P(A = (v, n)) = P(V = v)P(N = n),
\end{equation}
however this is na\"ive as verb-noun combinations aren't all as equally likely.
For example, it is much more probable to observe `cut onion' than `cut chopping board'.
In Long-term Feature Banks~\cite{wu2019_LongTermFeature}, the authors propose leveraging the prior knowledge of verb-noun co-occurrence in the training set $\mu(v, n)$ to weight the action prediction, \ie{}
\begin{equation}
P(A = (v, n)) \propto {\mu(v, n)}{P(V=v)}{P(N=n)}.
\label{eq:action-prior}
\end{equation}
The method in Eq.~\ref{eq:action-prior} does not allow zero-shot learning of unseen verb-noun combinations.
To remedy this, we apply Laplace smoothing to $\mu$ to avoid eliminating the possibility of recognising unseen actions.
We evaluate the relative benefit of using action priors in \cref{tab:action-prior}, finding it provides little benefit on the seen test set, but improves performance on the unseen test set by $\sim1\%$ point for top-1 accuracy.

\begin{table}[t]
  \midsepremove
  \small
  \centering
  \begin{tabular}{*{2}{Sl} *{4}{Sr}}
    \toprule
          &          & \multicolumn{2}{Sc}{Top-1} & \multicolumn{2}{Sc}{Top-5} \\
                     \cmidrule(lr){3-4}           \cmidrule(l){5-6}
    Model & Modality & \multicolumn{1}{Sc}{S1} & \multicolumn{1}{Sc}{S2} & \multicolumn{1}{Sc}{S1} & \multicolumn{1}{Sc}{S2} \\
    \midrule
    \multirow{2}{*}{TRN}   & RGB    & \ci{+0.05} & \ci{+1.33} & \ci{+0.14} & \ci{+1.43}\\
                           & Flow   & \ci{+0.01} & \ci{+1.43} & \cd{-0.50} & \ci{+0.75}\\
    \midrule
    \multirow{2}{*}{M-TRN} & RGB    & \cd{-0.14} & \ci{+0.99} & \ci{+0.70} & \ci{+2.80}\\
                           & Flow   & \cd{-0.25} & \ci{+0.68} & \cd{-0.61} & \ci{+0.24}\\
    \midrule
    \multirow{2}{*}{TSM}   & RGB    & \ci{+0.02} & \ci{+0.82} & \ci{+0.24} & \ci{+2.42}\\
                           & Flow   & \cd{-0.25} & \ci{+0.89} & \cd{-0.83} & \ci{+0.44}\\

    \bottomrule
  \end{tabular}
  \caption{Percentage point improvement on action task when using action prior
    across 8-segment ResNet-50 models.}
  \label{tab:action-prior}
  \midsepdefault
\end{table}

\section{Released Models}
All models required to reproduce the results in \cref{tab:backbone-comparison} are made available.
We release both RGB and flow models whose predictions can be combined to produce fusion results. To reproduce or compare to these results, the test set predictions should be submitted to the EPIC-Kitchens leaderboard\footnote{\url{https://epic-kitchens.github.io/2019\#challenges}} to calculate the performance.

The complexity of the models using ResNet-50 backbone is compared in \cref{tab:model-complexity},  

\begin{table}[t]
  \midsepremove
  \small
  \centering
  \begin{tabular}{SlSlSrSrSrSr}
    \toprule
          & \multicolumn{2}{Sc}{GFLOP/s} & \multicolumn{2}{Sc}{Params (M)} \\
          \cmidrule(lr){2-3}                \cmidrule(l){4-5}
    Model & RGB   & Flow                    & RGB   & Flow   \\
    \midrule
    TSN   & 33.12 & 35.33                   & 24.48 & 24.51 \\
    TRN   & 33.12 & 35.32                   & 25.33 & 25.35 \\
    M-TRN & 33.12 & 35.33                   & 27.18 & 27.21 \\
    TSM   & 33.12 & 35.33                   & 24.48 & 24.51 \\
    \bottomrule
  \end{tabular}
  \caption{Model parameter and FLOP/s count using a \mbox{ResNet-50} backbone with 8
    segments for a single video.}
  \label{tab:model-complexity}
  \midsepdefault
\end{table}

\section{Conclusion}

We have benchmarked 3 contemporary models for action recognition and analysed their performance, highlighting areas of good and poor performance.
TSM is competitive with M-TRN, and both outperform TSN.
These results highlight the necessity for temporal reasoning to recognise actions in EPIC-Kitchens.
Yet, the relatively low scores for top-1 accuracy show the challenge is far from solved.
Particular issues common to all models are the long-tailed nature of the dataset, fine-grained classes, and difficulty in generalising to unseen environments where we observe a significant drop across all metrics.

{\small
\bibliographystyle{ieee-fullname}
\bibliography{references}

\begin{thebibliography}{10}\itemsep=-1pt

\bibitem{damen2018_ScaleingEgocentricVision}
Dima Damen, Hazel Doughty, Giovanni~Maria Farinella, Sanja Fidler, Antonino
  Furnari, Evangelos Kazakos, Davide Moltisanti, Jonathan Munro, Toby Perrett,
  Will Price, and Michael Wray.
\newblock Scaling egocentric vision: The epic-kitchens dataset.
\newblock In {\em ECCV}, 2018.

\bibitem{goyal2017_SomethingSomethingVideo}
Raghav Goyal, Samira Ebrahimi~Kahou, Vincent Michalski, Joanna Materzynska,
  Susanne Westphal, Heuna Kim, Valentin Haenel, Ingo Fruend, Peter Yianilos,
  Moritz Mueller-Freitag, Florian Hoppe, Christian Thurau, Ingo Bax, and Roland
  Memisevic.
\newblock The "something something" video database for learning and evaluating
  visual common sense.
\newblock In {\em ICCV}, 2017.

\bibitem{he2016_DeepResidualLearning}
Kaiming He, Xiangyu Zhang, Shaoqing Ren, and Jian Sun.
\newblock Deep residual learning for image recognition.
\newblock In {\em CVPR}, 2016.

\bibitem{ioffe2015_BatchNormalizationAccelerating}
Sergey Ioffe and Christian Szegedy.
\newblock Batch normalization: Accelerating deep network training by reducing
  internal covariate shift.
\newblock In {\em ICML}, 2015.

\bibitem{kay2017_KineticsHumanAction}
Will Kay, Joao Carreira, Karen Simonyan, Brian Zhang, Chloe Hillier, Sudheendra
  Vijayanarasimhan, Fabio Viola, Tim Green, Trevor Back, Paul Natsev, Mustafa
  Suleyman, and Andrew Zisserman.
\newblock The kinetics human action video dataset, 2017.

\bibitem{kuehne2011_HMDBLargeVideo}
H. Kuehne, H. Jhuang, E. Garrote, T. Poggio, and T. Serre.
\newblock {HMDB}: a large video database for human motion recognition.
\newblock In {\em ICCV}, 2011.

\bibitem{lin2018_TSMTemporalShift}
J. Lin, Chuang. Gan, and S. Han.
\newblock Temporal shift module for efficient video understanding.
\newblock {\em arXiv}, 2018.

\bibitem{santoro2017_SimpleNeuralNetwork}
Adam Santoro, David Raposo, David~G Barrett, Mateusz Malinowski, Razvan
  Pascanu, Peter Battaglia, and Timothy Lillicrap.
\newblock A simple neural network module for relational reasoning.
\newblock In {\em NeurIPS}, 2017.

\bibitem{simonyan2014_TwoStreamConvolutional}
Karen Simonyan and Andrew Zisserman.
\newblock Two-stream convolutional networks for action recognition in videos.
\newblock In {\em NeurIPS}, 2014.

\bibitem{soomro2012_UCF101DatasetHuman}
Khurram Soomro, Amir~Roshan Zamir, and Mubarak Shah.
\newblock {UCF101}: A dataset of 101 human actions classes from videos in the
  wild, 2012.

\bibitem{szegedy2015_GoingDeeperConvolutions}
Christian Szegedy, Wei Liu, Yangqing Jia, Pierre Sermanet, Scott Reed, Dragomir
  Anguelov, Dumitru Erhan, Vincent Vanhoucke, and Andrew Rabinovich.
\newblock Going deeper with convolutions.
\newblock In {\em CVPR}, 2015.

\bibitem{wang2018_TemporalSegmentNetworks}
L. {Wang}, Y. {Xiong}, Z. {Wang}, Y. {Qiao}, D. {Lin}, X. {Tang}, and L. {Van
  Gool}.
\newblock Temporal segment networks for action recognition in videos.
\newblock {\em TPAMI}, 2019.

\bibitem{wu2019_LongTermFeature}
Chao{-}Yuan Wu, Christoph Feichtenhofer, Haoqi Fan, Kaiming He, Philipp
  Kr\"{a}henb\"{u}hl, and Ross Girshick.
\newblock {Long-Term Feature Banks for Detailed Video Understanding}.
\newblock In {\em {CVPR}}, 2019.

\bibitem{zhou2017_TemporalRelationalReasoning}
B. Zhou, A. Andonian, A. Oliva, and A. Torralba.
\newblock Temporal relational reasoning in videos.
\newblock In {\em ECCV}, 2018.

\end{thebibliography}
}

\end{document}